\documentclass{llncs}
\usepackage{makeidx}  
\usepackage{amssymb}
\usepackage{graphicx}
\usepackage{subfigure}
\usepackage{url}
\usepackage{xcolor}
\usepackage{multirow}
\usepackage[normalem]{ulem}
\begin{document}

\mainmatter  

\title{Hierarchical Gaussian Mixture Model with Objects Attached to Terminal and Non-terminal Dendrogram Nodes}

\titlerunning{Hierarchical GMM with Objects Attached to All Dendrogram Nodes}

%
%
\author{{\L}ukasz P. Olech \and Mariusz Paradowski}
\authorrunning{{\L}ukasz P. Olech \and Mariusz Paradowski}
\tocauthor{{\L}ukasz P. Olech, Mariusz Paradowski}

\institute{Department of Computational Intelligence\\Wroclaw University of Technology, Poland\\ \email{lukasz.olech@pwr.edu.pl}}

%
%

\maketitle

\begin{abstract}
A hierarchical clustering algorithm based on Gaussian mixture model is presented.
The key difference to regular hierarchical mixture models is the ability to store objects in both terminal and non-terminal nodes.
Upper levels of the hierarchy contain sparsely distributed objects, while lower levels contain densely represented ones. 
As it was shown by experiments, this ability helps in noise detection (modelling).
Furthermore, compared to regular hierarchical mixture model, the presented method generates more compact dendrograms with higher quality measured by adopted F-measure.
\end{abstract}

\section{Introduction}

This paper addresses the topic of {\em hierarchical data clustering} which is a countertype to {\em flat clustering}.
Flat clustering approaches generate groups without structural connections between them.
Hierarchical clustering algorithms generate groups and arrange them in a tree structured manner.
In such a tree structure (known as a {\em dendrogram}) all child clusters are attached to their parent cluster.
Clusters without any further children are called {\em terminal nodes} or {\em tree leaves}.
Clusters with attached child clusters are called {\em non-terminal} or {\em internal nodes}.

Hierarchical clustering algorithms can be divided into two categories depending on the objects attachment to the generated groups.
The first category represents methods that attach objects only to {\em terminal nodes}, and {\em non-terminal} nodes of the hierarchy remain empty.
This kind of methods are the majority of hierarchical data clustering methods.
It is possible to fill an internal node with objects, by gathering all objects belonging to its child nodes.
The second category represents methods attaching objects both to {\em internal nodes} and {\em tree leaves}.
All tree leaves need to have at least one attached object.
Internal nodes can have attached objects or remain empty.
The {\em key difference} is that if an object is attached to the internal node, it is {\em not attached} to any of its child nodes.
Methods belonging to this category are the minority.
The presented research addresses this category.

This paper is organized as follows. In following subsections we give the necessary background of clustering problems. The proposed method is introduced in the second section. The third section presents the experimental results and comparison with regular hierarchical Gaussian mixture model. Finally, the fourth section summarises this paper.

\subsection{Hierarchical approaches to clustering}

One of the earliest approaches of hierarchical clustering is {\em hierarchical agglomerative clustering} (HAC) \cite{murtagh-1983}.
HAC creates a dendrogram with all objects attached to its leaves.
At each level of the hierarchy, two groups are merged.
As a result, the created structure is an {\em unbalanced binary tree}.
Various merging schemes are available, e.g., {\em Ward criterion}~\cite{ward-1963}, {\em single-link}~\cite{sibson-1973} or {\em complete-link}~\cite{defays-1977}.

Both binary and non-binary hierarchies can be constructed using various extensions of the {\em k-means} algorithm~\cite{hartigan-1979,jain-2010}.
Usage of hierarchical k-means leads to two major consequences comparing to flat k-means.
First, the clustering process is {\em much faster} because the number of groups in a tree path is much lower.
This is especially important if the number of clusters and the volume of data are high.
Second, the overall quality of clustering tends to be {\em worse}, because cluster centers are not optimized simultaneously.
One of the key problems of k-means clustering (both flat and hierarchical) is estimation of the number of clusters.
There are many attempts to address this issue, e.g. {\em x-means} algorithm~\cite{dan-2000}.

Hierarchical clustering using a probabilistic approach is also possible, e.g.,~\cite{liu-2002}.
The milestone in probabilistic clustering was the formulation of the {\em expectation maximization} (EM) algorithm~\cite{dempster-1977}.
Hierarchical setup of mixture models can be trained using modified EM~\cite{carneiro-2007}.
One of the most common choices for mixture components is multivariate normal distribution.

\subsection{Clustering in the presence of noise}

Yet another important issue is clustering of data in the presence of {\em noise} or {\em outliers}.
There are two common solutions to this problem.
The first solution consists of two stages, e.g.,~\cite{byers-1998}.
In the initial stage data is filtered in order to detect and remove outliers.
Then in the second stage clusterisation is performed only on the accepted data.
The second solution is to directly incorporate the noise model into the clustering process.
Usually, the type or distribution of noise or outliers is not known.
Various assumptions regarding these distributions have to be made.
Exemplary, DBSCAN~\cite{ester-1996} and OPTICS~\cite{ankerst-1999} clustering algorithms assume a minimum density of the meaningful data.
In probabilistic clustering, noise can be directly modeled by appropriate mixture components, e.g.,~\cite{campbell-1997,fraley-2002}.

\subsection{Problem formulation, motivation and contribution}

Probabilistic approach to clustering can be formulated using the parametric model.
The key issue is the formulation of an appropriate {\em probability density function} (PDF).
There are several forms of the probabilistic density function.
{\em Gaussian mixture model} is one of the most prominent~\cite{fraley-2002}.
Let the {\em Gaussian mixture model} $G$ with $n$ mixture components be defined as:
\begin{equation}
G(w, \mu, \Sigma) = \sum_{i=1}^{n}w_iN(\mu_i, \Sigma_i), \quad w_i \in {\langle}0, 1\rangle, \quad \sum_{i=1}^{n} w_i = 1,
\label{gaussian-mixture}
\end{equation}
and $N(\mu, \Sigma)$ represents the multivariate normal distribution, $w = [w_1,...,w_n]$, $\mu = [\mu_1, ..., \mu_n]$, $\Sigma = [\Sigma_1,...,\Sigma_n]$.
In such case clustering problem becomes a probability density function estimation problem, where PDF parameters maximize {\em likelihood} $\mathcal{L}$:
\begin{equation}
\left\langle w^*, \mu^*, \Sigma^* \right\rangle = \arg\max_{\left\langle w, \mu, \Sigma \right\rangle} \mathcal{L}(w, \mu, \Sigma | x_1,...,x_m),
\label{likelihood}
\end{equation}
where: $x_1,...,x_m$ are the data vectors. This is typically solved by the EM algorithm, but other methods are also available, e.g.,~\cite{figueiredo-2002,verbeek-2003,zivkovic-2004}.

Gaussian mixture model fits to data distributed among several clusters, but does not model {\em outliers}~\cite{fraley-2002}.
Data not fitting to the assumed distribution can be interpreted in several ways, including: {\em noise}, {\em measurement errors} or {\em sparser representation of meaningful objects}.
Statistical modelling of noisy data requires making assumptions on the noise distribution.
The data distribution is usually combined with noise distribution, e.g.,~\cite{minka-2001}.

In the presented approach we follow the third interpretation of the not fitting data, i.e., sparser representation of meaningful objects.
We do not want to reject the data, we want to model it on some level of the generated hierarchy.
Data bound to the parent clusters should have lesser density comparing to the data bound to the child clusters.
In the paper we show a simple approach to adapt hierarchical Gaussian mixture model to handle objects attached to any node in the tree.
Similar to noise modelling~\cite{campbell-1997,fraley-2002} we add an additional mixture component to the mixture model.
But unlike that approaches, we do not estimate it, but {\em directly take it from the higher level of the hierarchy}.
As a consequence, parameters of the adapted mixture model are estimated in an identical manner as for the classic mixture model.
They can be estimated both using EM or any other appropriate approach.

\section{Proposed approach}

The proposed approach is an extension of a hierarchical setup of Gaussian mixture models.
At each level of the hierarchy an additional mixture component, called {\em background component}, is introduced.
This component is responsible for capturing outliers at a given level.
Unlike all other mixture components, it is not estimated, but directly inherited from the higher level of the hierarchy.
Root level also has this additional component.
Its parameters are estimated (by definition) from all available data.

\subsection{Formal model of the hierarchy}
Let us define the model of the hierarchy in a {\em recursive} way.
Any {\em parent node} has all its {\em child nodes}.
A tree node $T$ generated from a data set $X$ is defined as:
\begin{equation}
\label{tree-node-reccursive-form}
T(X) : {\langle}n, G_B, B \subseteq X, [T_1(X_1),...,T_n(X_n)]{\rangle},
\end{equation}
where: 
\begin{equation}
B \cup \bigcup_{i = 1}^{n}{X_i} = X, \quad \forall_{i \in [1, n]} B \cap {X_i} = \emptyset, \quad \forall_{i, j \in [i, n]} i \neq j \Rightarrow X_i \cap X_j = \emptyset
\end{equation}
and: $n$ is the maximum number of child nodes (and mixture components),
$G_B$ is the {\em Gaussian mixture model with background component} $N(\mu_B, \Sigma_B)$:
\begin{eqnarray}
G_B(\alpha, w, \mu_B, \mu, \Sigma_B, \Sigma) = \alpha N(\mu_B, \Sigma_B) + (1 - \alpha) G(w, \mu, \Sigma) = \nonumber \\
=\alpha N(\mu_B, \Sigma_B) + \sum_{i=1}^{n}(1 - \alpha)w_iN(\mu_i, \Sigma_i), \\
\alpha \in {\langle}0,1{\rangle}, {\quad} \mu_B = E[X], {\quad} \Sigma_B = Var[X],
\label{background-gaussian-mixture}
\end{eqnarray}
$B \subseteq X$ is the data subset attached to the node $T$, related to background mixture component $N(\mu_B, \Sigma_B)$,
$T_1, ..., T_n$ are child nodes or {\em void}. Mixture component $G_B$ and set $B$ are representing the data that remain in tree node $T$.
They are the key difference when comparing to classic hierarchical clustering methods.

\subsection{Hierarchy generation}

Cluster hierarchy generation is done in an recursive way.
First, the top level is generated and its parameters are estimated.
Later on, child levels are added sequentially in {\em breadth-first} manner.
For each level the process terminates if a stop criterion is reached.
This process is similar to the one used in hierarchical k-means approach~\cite{steinbach-2000}.
It allows a dynamic generation of the hierarchical structure.

As shown in the formal model, each level of the hierarchy contains only a subset of the data.
The top level starts with all the data.
Expectation maximization method is used to estimate the Gaussian mixture model.
Because the proposed method is iterative, stochastic, and strongly depended on cluster initialization, several cluster reinitialisations should be performed.
Thus the number of cluster reinitialization $R$ and number of EM iterations $N$ are the parameters.

Clusters initialization is based on choosing random $n$ distinct points from the data and set them as initial centres $\mu$ of new clusters. Initial covariances $\Sigma$ of that clusters are the same as parent cluster covariance. 
Full covariance matrices are used.
When covariance matrix is non-invertible, regularization is introduced.
Mixing coefficients (see eq. \ref{background-gaussian-mixture}) are initialized as equal values:
\begin{equation}
\label{initial-mixing-coef}
\alpha = \frac{1}{n+1}, \quad (1 - \alpha)w_i = \frac{1}{n+1}.
\end{equation}
The denominator takes into account $n$ newly created clusters and a background cluster.
The data is distributed to all mixture components, according to data probability assignments.
A single data instance is assigned to the mixture component with highest probability of generating that instance.
As a result, some mixture components, including the background component, may remain empty.
After initialisation, the EM algorithm works through $N$ iterations, changing initial values of $\mu, \Sigma, w$ and $\alpha$.
After performing $R$ reinitialisations, a solution with the largest likelihood is chosen (see eq. \ref{likelihood}) as the final one.

All mixture components with assigned data instances generate child nodes.
The above process repeats for every generated node.
In case a mixture component does not receive any data, it also does not generate a child node.
The child nodes generation process is terminated when a stop criterion is reached.
There are two stop criteria and each of them terminates the method.
The first stop criterion is connected with the content of current leaf nodes.
The clustering process proceeds only on those leaf nodes that contain at least $k$ different data samples.
The algorithm terminates when there are no leaf nodes to split or all data is assigned to background mixture component $B$.
The second criterion occurs when provided $W$ overall number of nodes was created.

\section{Experimental verification}

Experimental verification of the proposed approach consists of two parts.
In the first part we give illustrative examples to demonstrate the idea behind the method.
Manually prepared toy datasets are used for visualization purposes.
In the second part we test the proposed approach on a set of benchmark datasets from UCI repository~\cite{uci-2013}.
We choose well-known \textit{iris, wine, glass identification} and \textit{image segmentation} datasets varying in number of classes, attributes and instances, as shown in Table \ref{uci-dataset-stats}.

\begin{table}[h!tb]
\caption{Original (without additional noise) UCI dataset statistics.}
\begin{center}
\begin{tabular}{c|c|c|c}
dataset name & instances & attributes & classes \\ 
\hline
iris & 150 & 4 & 3 \\
wine & 178 & 13 & 3 \\
glass identification & 214 & 9 & 6 \\
image segmentation & 2100 & 19 & 7 \\
\end{tabular}
\end{center}
\label{uci-dataset-stats}
\end{table}
  
Since the mentioned datasets do not contain any noise points we added them manually. Noise points are uniformly distributed among original points. 
In each dataset, the number of noise points is equal to the half of the number of original points.
The proposed approach is compared to a standard hierarchical set-up of Gaussian mixture model.

In order to compare the obtained results on the benchmark datasets we use a metric based on {\em F-measure}~\cite{larsen-1999}. 
It takes a class attribute into consideration and yield a grouping quality by considering the whole dendrogram, not only a chosen level. 
This makes the measure adequate for hierarchical methods.
\textit{F-measure} is calculated for each generated group $B$ with respect to each class $C$:
\begin{equation}
P(X_i, C_c) = \frac{N_{ic}}{|X_i|},\quad R(X_i, C_c) = \frac{N_{ic}}{N_{C_c}},
\end{equation}
\begin{equation}
F_{ic} = \frac{2P(X_i, C_c)R(X_i, C_c)}{P(X_i, C_c)+R(X_i, C_c)},
\end{equation}
where:
$F_{ic}$ -- \textit{F-measure} for \textit{i-th} group and \textit{c-th} class,
$P(X_i, C_c)$ -- precision and
$R(X_i, C_c)$ -- recall, for \textit{i-th} group with respect to \textit{c-th class},
$N_{ic}$ is the number of objects from \textit{c-th} class which are within \textit{i-th} group,
$N_{C_c}$ is the number of object from \textit{c-th} class in the entire tree and 
$|X_i|$ is the number of objects that are within \textit{i-th} cluster.
Noise points are not regarded as an additional class, they are only counted in each $|X_i|$.
Given the above definitions, F-measure for a chosen class $C_c$ is defined as the maximum value of the measure over all nodes of the tree:
\begin{equation}
\label{f-measure-for-each-class}
F(C_c) = \max_{i} F_{ic}.
\end{equation}
Finally, it is averaged over all classes giving F-Measure for whole hierarchy:
\begin{equation}
\label{hierarchy-f-measure}
F = \frac{1}{N}\sum_{c=1}^{|C|}N_{C_c}F(C_c),
\end{equation}
where: 
$|C|$ is number of classes used in dataset, 
$N$ is the total number of objects (including noise points) and 
$N_{C_c}$ is the number of data objects of class $c$.
Proposed evaluation criterion has the ability to explore hierarchy structure, which is a key point in the proposed method.
$F$ maximum value is 1 and minimum is 0.
Better hierarchies have higher $F$ values.

\subsection{Manually generated data with noise -- an illustration}

All results presented in this section are two dimensional toy examples.
Their sole purpose is to illustrate the behavior of the proposed method.
The following examples are presented:
\begin{enumerate}
\item three groups with a large central group and a small amount of noise ({\em LC}),
\item small circular data clusters with a small amount of noise ({\em LN}),
\item small circular data clusters with a large amount of noise ({\em HN}).
\end{enumerate}
\begin{figure}[h!tb]
\begin{center}
\subfigure[large center (LC)]{\fbox{\includegraphics[width=3.05cm]{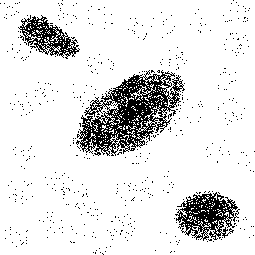}}}
\subfigure[LC, level 1]{\fbox{\includegraphics[width=3.05cm]{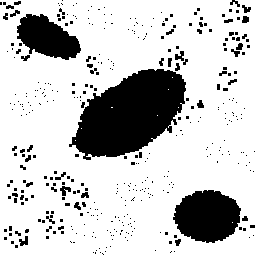}}}
\subfigure[LC, level 2]{\fbox{\includegraphics[width=3.05cm]{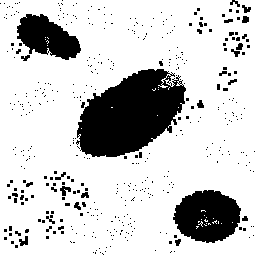}}}\\
\subfigure[low noise (LN)]{\fbox{\includegraphics[width=2.7cm]{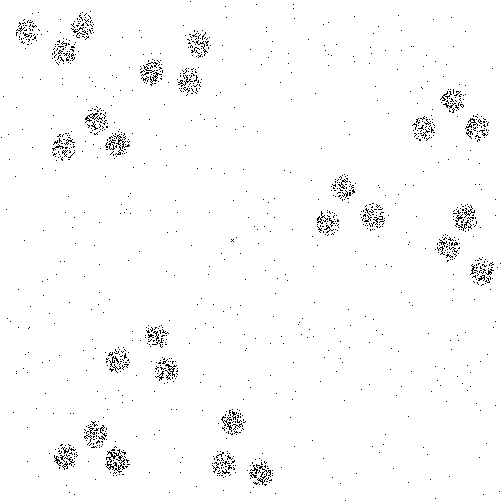}}}
\subfigure[LN, level 1]{\fbox{\includegraphics[width=2.7cm]{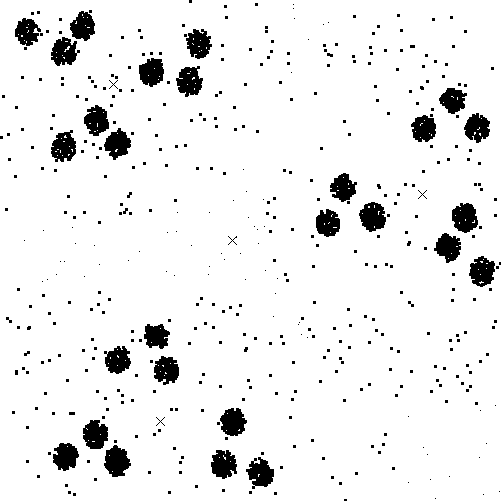}}}
\subfigure[LN, level 2]{\fbox{\includegraphics[width=2.7cm]{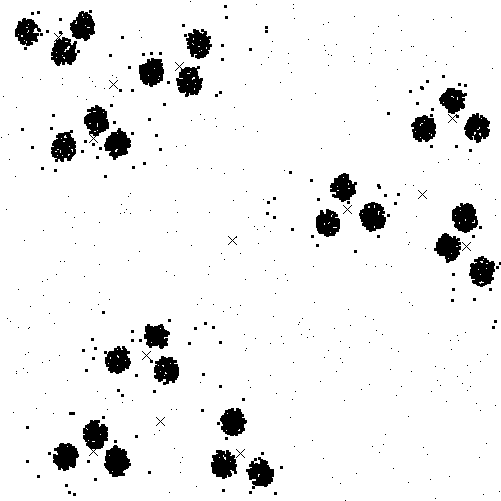}}}
\subfigure[LN, level 3]{\fbox{\includegraphics[width=2.7cm]{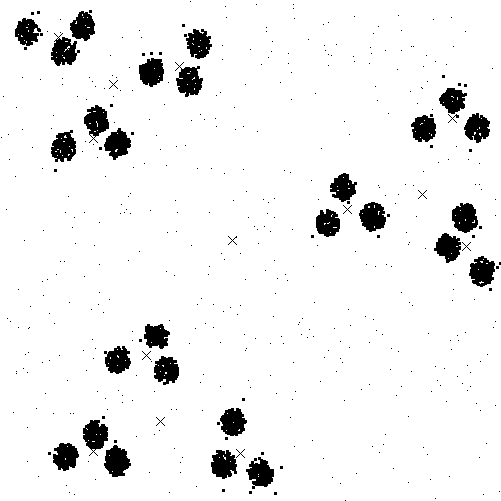}}} \\
\subfigure[high noise (HN)]{\fbox{\includegraphics[width=2.7cm]{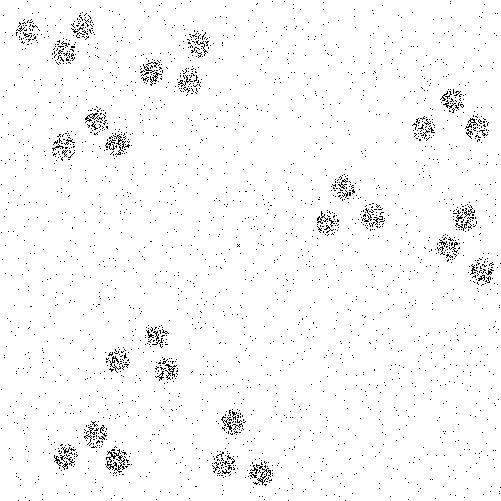}}}
\subfigure[HN, level 1]{\fbox{\includegraphics[width=2.7cm]{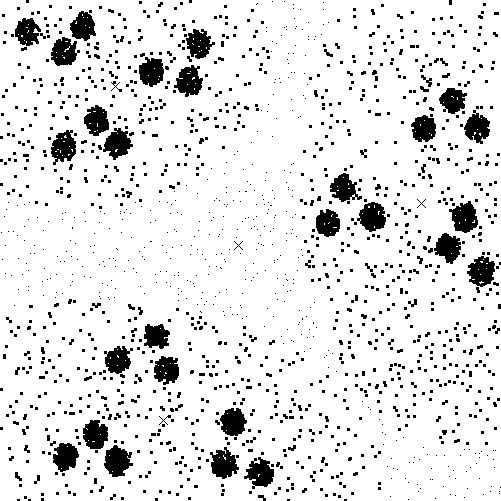}}}
\subfigure[HN, level 2]{\fbox{\includegraphics[width=2.7cm]{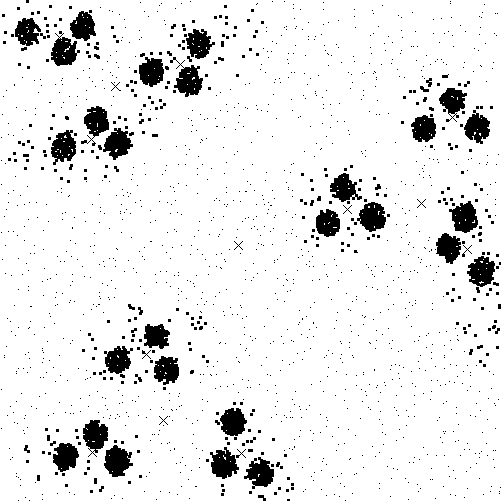}}}
\subfigure[HN, level 3]{\fbox{\includegraphics[width=2.7cm]{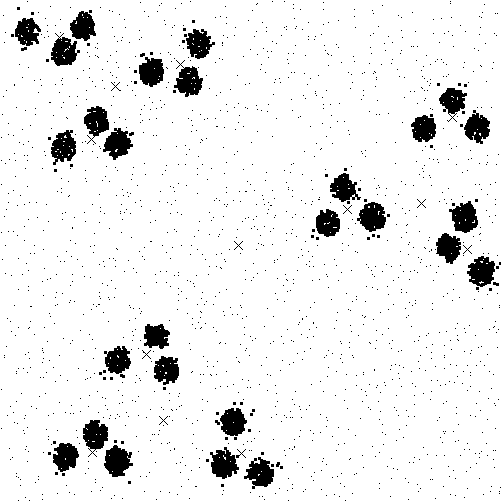}}}
\end{center}
\caption{Two dimensional toy datasets with a various amount of noise. First column shows the data points. Second, third and fourth columns show the clustering results at different levels of the hierarchy. Data attached to proposed background model are shown as small pixels, data attached to the mixture model are shown as large pixels.}
\label{generated-data-noise}
\end{figure}
Both the data and clustering results for the toy datasets are shown in Fig. \ref{generated-data-noise}.
The method has some ability to capture less dense data.
This data is attached to the intermediate nodes of the hierarchy.
The additional background model component captures these instances.
In consequence they are automatically bound to the node related to the background component.
At the same time, densely distributed data is moved to the bottom of the hierarchy.
This can be observed (to some extent) at all presented test cases.

\subsection{UCI benchmark datasets}

The second part of the experiments addresses the clustering of the UCI benchmark datasets.
Instances of all processed datasets have both feature vectors and class assignment.
Feature vectors without class information are used in the clustering process.
Available class assignment is used in the evaluation process.

Two methods are compared: (1) the proposed Gaussian mixture model with outlier modelling and (2) classic Gaussian mixture model. 
The first method is denoted as \textit{B} and the second as \textit{G}.
Both methods are trained using the same expectation-maximization routine.
Hierarchies of both models are constructed in the same manner.
Two quality estimates are shown: (1) log-likelihood values to address data fitting to the distribution, (2) f-measure values to check if the generated groups are meaningful.

\begin{table}[h!tb]
\caption{Comparison of the proposed model ($\mu_{B}$, $\sigma_{B}$) with the reference Gaussian mixture model ($\mu_{G}$, $\sigma_{G}$). Both log-likelihood values are F-measure values are shown. Higher F-measure values are marked in bold.}
\begin{center}
\begin{tabular}{c|c|rr|rr|rr|rr|cc}
dataset 					& no of 	& \multicolumn{4}{c|}{log-likelihood} & \multicolumn{4}{c|}{f-measure} & \multicolumn{2}{c}{significance test} \\ 
name    					& nodes 	& $\mu_{B}$ & $\sigma_{B}$ & $\mu_{G}$ & $\sigma_{G}$ & $\mu_{B}$ & $\sigma_{B}$ & $\mu_{G}$ & $\sigma_{G}$ & $U$ & winner \\ 
\hline
\multirow{10}{*}{iris}	
    						& 2  	& $247$ & ${7}$  & $107$ & ${0}$  & $\bf 0.59$ & ${0.045}$ & $0.36$ & ${0.000}$     & 0 & B \\
						& 3  	& $359$ & ${0}$  & $288$ & ${0}$  & $\bf 0.75$ & ${0.002}$ & $0.62$ & ${0.000}$     & 0 & B \\
						& 4 		& $359$ & ${1}$  & $355$ & ${14}$ & $\bf 0.75$ & ${0.003}$ & $0.62$ & ${0.000}$     & 0 & B \\
						& 5 		& $490$ & ${24}$ & $429$ & ${37}$ & $\bf 0.77$ & ${0.013}$ & $0.67$ & ${0.061}$     & 0 & B \\
						& 6 		& $483$ & ${27}$ & $503$ & ${53}$ & $\bf 0.77$ & ${0.081}$ & $0.69$ & ${0.062}$     & 830 & B \\
						& 7 		& $480$ & ${28}$ & $584$ & ${5}$  & $\bf 0.77$ & ${0.058}$ & $0.75$ & ${0.001}$     & 1385 & B \\
						& 8 		& $601$ & ${29}$ & $623$ & ${23}$ & $\bf 0.79$ & ${0.044}$ & $0.74$ & ${0.006}$     & 697 & B \\
						& 9 		& $629$ & ${24}$ & $642$ & ${29}$ & $\bf 0.80$ & ${0.046}$ & $\bf 0.80$ & ${0.071}$ & 3760 & B \\
						& 10 	& $614$ & ${19}$ & $680$ & ${43}$ & $\bf 0.80$ & ${0.046}$ & $0.78$ & ${0.070}$ 	& 2679 & B \\
\hline
\multirow{10}{*}{wine}	
    						& 2  	& $366$ & ${1}$  & $307$ & ${0}$  & $\bf 0.40$ & ${0.002}$ & $0.37$ & ${0.000}$     & 0 & B \\
						& 3  	& $402$ & ${2}$  & $392$ & ${1}$  & $\bf 0.42$ & ${0.005}$ & $0.41$ & ${0.005}$     & 2033 & B \\
						& 4 		& $403$ & ${2}$  & $445$ & ${21}$ & $\bf 0.42$ & ${0.003}$ & $0.41$ & ${0.005}$     & 1539 & B \\
						& 5 		& $466$ & ${8}$ & $478$ & ${20}$ & $\bf 0.42$ & ${0.005}$ & $\bf 0.42$ & ${0.017}$  & 5298 & --\\
						& 6 		& $466$ & ${11}$ & $529$ & ${8}$ & $0.42$ & ${0.005}$ & $\bf 0.43$ & ${0.016}$      & 6188 & G \\
						& 7 		& $470$ & ${8}$ & $561$ & ${8}$  & $0.42$ & ${0.004}$ & $\bf 0.44$ & ${0.005}$      & 10000 & G\\
						& 8 		& $527$ & ${19}$ & $588$ & ${23}$ & $0.42$ & ${0.010}$ & $\bf 0.44$ & ${0.006}$     & 9372 & G \\
						& 9 		& $524$ & ${20}$ & $599$ & ${20}$ & $0.42$ & ${0.008}$ & $\bf 0.44$ & ${0.008}$ & 9766 & G \\
						& 10 	& $531$ & ${20}$ & $617$ & ${31}$ & $0.42$ & ${0.008}$ & $\bf 0.44$ & ${0.010}$ 	& 9753 & G \\
\hline
\multirow{10}{*}{glass}	
    						& 2  & $809$   & ${1}$  & $338$ & ${0}$  & $\bf 0.41$ & ${0.000}$ & $0.29$ & ${0.000}$     & -- & B \\
						& 3  & $1121$ & ${1}$  & $1121$ & ${0}$  & $\bf 0.40$ & ${0.001}$ & $\bf 0.40$ & ${0.000}$   & 4950 & B \\
						& 4 		& $1275$  & ${28}$  & $1227$ & ${15}$ & $\bf 0.45$ & ${0.045}$ & $0.40$ & ${0.000}$  & 600 & B \\
						& 5 	& $1299$ & ${48}$ & $1274$ & ${36}$ & $\bf 0.46$ & ${0.044}$ & $0.45$ & ${0.050}$  & 5084 & -- \\
						& 6 	& $1314$ & ${49}$ & $1372$ & ${19}$ & $0.44$ & ${0.043}$ & $\bf 0.46$ & ${0.054}$      & 5537 & -- \\
						& 7 	& $1469$ & ${23}$ & $1432$ & ${5}$  & $\bf 0.50$ & ${0.008}$ & $\bf 0.50$ & ${0.027}$   & 8715 & G \\
						& 8 	& $1477$ & ${19}$ & $1465$ & ${15}$ & $0.49$ & ${0.022}$ & $\bf 0.50$ & ${0.038}$     & 6476 & G \\
						& 9 	& $1495$ & ${5}$ & $1501$ & ${19}$ & $0.49$ & ${0.010}$ & $\bf 0.51$ & ${0.015}$ & 9032 & G \\
						& 10 	& $1532$ & ${14}$ & $1533$ & ${16}$ & $\bf 0.50$ & ${0.013}$ & $\bf 0.50$ & ${0.022}$ & 7632 & G \\
\hline
\multirow{10}{*}{segmentation}	
    						& 2  & $-6605$ & ${11}$  & $-6678$ & ${0}$  & $\bf 0.29$ & ${0.012}$ & $ 0.28$ & ${0.000}$     & -- & -- \\
						& 3  & $-6466$ & ${3}$  & $-6481$ & ${1}$  & $\bf 0.50$ & ${0.000}$ & $\bf 0.50$ & ${0.001}$   & 495 & B \\
						& 4 	 & $-6466$ & ${3}$ & $-5413$ & ${154}$ & $\bf 0.50$ & ${0.000}$ & $\bf 0.50$ & ${0.000}$  & 600 & -- \\
						& 5 	 & $-5348$ & ${76}$ & $-5406$ & ${268}$ & $ 0.50$ & ${0.001}$ & $\bf 0.52$ & ${0.042}$  & 5084 & -- \\
						& 6  & $-5643$ & ${1001}$ & $-4524$ & ${72}$ & $ 0.49$ & ${0.017}$ & $\bf 0.54$ & ${0.042}$      & 5537 & -- \\
						& 7 	 & $-5470$ & ${710}$ & $-4483$ & ${115}$  & $ 0.50$ & ${0.014}$ & $\bf 0.56$ & ${0.038}$   & 8715 & G \\
						& 8  & $-5257$ & ${715}$ & $-4388$ & ${366}$ & $0.50$ & ${0.019}$ & $\bf 0.56$ & ${0.038}$     & 6476 & G \\
						& 9 	 & $-5342$ & ${763}$ & $-4253$ & ${359}$ & $0.50$ & ${0.026}$ & $\bf 0.57$ & ${0.049}$ & 9032 & G \\
						& 10 & $-5257$ & ${850}$ & $-3959$ & ${278}$ & $ 0.50$ & ${0.018}$ & $\bf 0.58$ & ${0.050}$ & 7632 & G \\
\end{tabular}
\end{center}
\label{uci-reference-results}
\end{table}

Performed experiments consider mentioned quality estimators when $W$ parameter vary between $2$ and $10$.
In all conducted experiments we set \textit{n} parameter as a constant equal to $2$.
First of all, we found the best parameters configuration ($N$ and $R$) for each method per single dataset instance and $W$ value.
Then, because of stochastic nature of both methods, we have performed 100 trials for each of dataset and $W$ parameter value, calculating mean value $\mu$ and sample standard deviation $\sigma$.
Moreover we conducted the \textit{Wilcoxon rank-sum test}~\cite{mann-1947} on calculated F-measure in order to show the statistic significance of the obtained results.
Statistic value $U$ is calculated with alpha level ($\alpha$) equal to $0.05$. Null hypotheses $H_0$ are equality of population distributions and alternative hypotheses $H_A$ may vary (depending on the corresponding F-measure $\mu$ values).
When F-measure mean values $\mu$ were different, then we performed a one-tailed test whereas equal means results in two-tailed. 
Achieved results are shown in Tab. \ref{uci-reference-results}. In that table the \textit{winner} column shows whether there is statistical evidence to reject the null hypothesis and assume an alternative one. 

Experiments results in Tab. \ref{uci-reference-results} shows that the proposed background component improves the quality of generated dendrograms, when considering data class labels. This is especially visible when maximum number of nodes \textit{W} is less than 5. Our method, though has the ability of creating compact dendrograms with better quality than the method without the background component. It is desired, because shorter trees have better generalisation abilities. Moreover, considering the \textit{iris} dataset, the background component helps obtaining higher F-measure in all cases, comparing to regular hierarchical Gaussian mixture model. Proposed method reaches statistically higher average F-measure results in 16 cases whereas the regular method wins only 13 times. There have been 7 draws. 

\section{Summary}

A hierarchical grouping method is presented.
It has the ability to attach objects both to terminal and non-terminal nodes.
It is an extension of the classic Gaussian mixture model.
The mixture is extended with an additional component responsible for outlier modelling.
Parameters of this mixture component are not estimated, but directly inherited from higher levels of the hierarchy.

Conducted experiments show that the proposed modification allows to treat part of the data as sparser representation of meaningful objects. Though upper levels of hierarchy consist of sparsely distributes data. This can be used in noise or outliers modelling. Comparison between regular hierarchic GMM and hierarchic GMM with proposed modification shows that the background component helps to improve the quality of short hierarchies in real datasets with random noise.

\bibliographystyle{splncs03}
\bibliography{bibliography}

\end{document}